\documentclass[letterpaper, 10 pt, journal, twoside]{IEEEtran}  % Comment this line out if you need a4paper

\ifCLASSINFOpdf
\else
\fi

\usepackage{graphics}
\usepackage{graphicx}
\usepackage{booktabs}
\usepackage{multirow}

\begin{document}
%
% paper title
% Titles are generally capitalized except for words such as a, an, and, as,
% at, but, by, for, in, nor, of, on, or, the, to and up, which are usually
% not capitalized unless they are the first or last word of the title.
% Linebreaks \\ can be used within to get better formatting as desired.
% Do not put math or special symbols in the title.
\title{Graph-Guided Deformation for Point Cloud Completion}
%
%
% author names and IEEE memberships
% note positions of commas and nonbreaking spaces ( ~ ) LaTeX will not break
% a structure at a ~ so this keeps an author's name from being broken across
% two lines.
% use \thanks{} to gain access to the first footnote area
% a separate \thanks must be used for each paragraph as LaTeX2e's \thanks
% was not built to handle multiple paragraphs
%

% \author{Michael~Shell,~\IEEEmembership{Member,~IEEE,}
%         John~Doe,~\IEEEmembership{Fellow,~OSA,}
%         and~Jane~Doe,~\IEEEmembership{Life~Fellow,~IEEE}% <-this % stops a space
% \thanks{M. Shell was with the Department
% of Electrical and Computer Engineering, Georgia Institute of Technology, Atlanta,
% GA, 30332 USA e-mail: (see http://www.michaelshell.org/contact.html).}% <-this % stops a space
% \thanks{J. Doe and J. Doe are with Anonymous University.}% <-this % stops a space
% \thanks{Manuscript received April 19, 2005; revised August 26, 2015.}}

\author{Jieqi Shi$^{1}$, Lingyun Xu$^{2}$, Liang Heng$^{3}$ and Shaojie Shen$^{4}$%
\thanks{Manuscript received: February 24, 2021; Revised April 30, 2021; Accepted July 5, 2021. This paper was recommended for publication by Editor Cesar Cadena upon evaluation of the Associate Editor and Reviewers' comments. This work was supported by The Research Grants Council General Research Fund (RGC GRF) project 16213717, HKUST Postgraduate Studentship. $^{1}$Jieqi and $^{4}$Shaojie Shen are with Department of Electronic and Computer
Engineering, Hong Kong University of Science and Technology
        {\tt\footnotesize $\{$jshias, eeshaojie$\}$@connect.ust.hk} $^{2}$Lingyun Xu and $^{3}$Liang Heng are with Dji Co.
        {\tt\footnotesize $\{$judy.xu, Liang.Heng$\}$@dji.com} Digital Object Identifier (DOI): see top of this page.}
}
% note the % following the last \IEEEmembership and also \thanks - 
% these prevent an unwanted space from occurring between the last author name
% and the end of the author line. i.e., if you had this:
% 
% \author{....lastname \thanks{...} \thanks{...} }
%                     ^------------^------------^----Do not want these spaces!
%
% a space would be appended to the last name and could cause every name on that
% line to be shifted left slightly. This is one of those "LaTeX things". For
% instance, "\textbf{A} \textbf{B}" will typeset as "A B" not "AB". To get
% "AB" then you have to do: "\textbf{A}\textbf{B}"
% \thanks is no different in this regard, so shield the last } of each \thanks
% that ends a line with a % and do not let a space in before the next \thanks.
% Spaces after \IEEEmembership other than the last one are OK (and needed) as
% you are supposed to have spaces between the names. For what it is worth,
% this is a minor point as most people would not even notice if the said evil
% space somehow managed to creep in.

% The paper headers
%\markboth{Journal of \LaTeX\ Class Files,~Vol.~14, No.~8, August~2015}%
%{Shell \MakeLowercase{\textit{et al.}}: Bare Demo of IEEEtran.cls for IEEE Journals}
\markboth{IEEE Robotics and Automation Letters. Preprint Version. Accepted July, 2021}
{Shi \MakeLowercase{\textit{et al.}}: Graph-Guided Deformation for Point Cloud Completion} 

% The only time the second header will appear is for the odd numbered pages
% after the title page when using the twoside option.
% 
% *** Note that you probably will NOT want to include the author's ***
% *** name in the headers of peer review papers.                   ***
% You can use \ifCLASSOPTIONpeerreview for conditional compilation here if
% you desire.

% If you want to put a publisher's ID mark on the page you can do it like
% this:
%\IEEEpubid{0000--0000/00\$00.00~\copyright~2015 IEEE}
% Remember, if you use this you must call \IEEEpubidadjcol in the second
% column for its text to clear the IEEEpubid mark.

% use for special paper notices
%\IEEEspecialpapernotice{(Invited Paper)}

% make the title area
\maketitle

% As a general rule, do not put math, special symbols or citations
% in the abstract or keywords.
\begin{abstract}
For a long time, the point cloud completion task has been regarded as a pure generation task. After obtaining the global shape code through the encoder, a complete point cloud is generated using the shape priorly learnt by the networks. However, such models are undesirably biased towards prior average objects and inherently limited to fit geometry details. In this paper, we propose a Graph-Guided Deformation Network, which respectively regards the input data and intermediate generation as controlling and supporting points, and models the optimization guided by a graph convolutional network(GCN) for the point cloud completion task. Our key insight is to simulate the least square Laplacian deformation process via mesh deformation methods, which brings adaptivity for modeling variation in geometry details. By this means, we also reduce the gap between the completion task and the mesh deformation algorithms. As far as we know, we are the first to refine the point cloud completion task by mimicing traditional graphics algorithms with GCN-guided deformation. We have conducted extensive experiments on both the simulated indoor dataset ShapeNet, outdoor dataset KITTI, and our self-collected autonomous driving dataset Pandar40. The results show that our method outperforms the existing state-of-the-art algorithms in the 3D point cloud completion task.
\end{abstract}

% Note that keywords are not normally used for peerreview papers.
% \begin{IEEEkeywords}
% IEEE, IEEEtran, journal, \LaTeX, paper, template.
% \end{IEEEkeywords}

% For peer review papers, you can put extra information on the cover
% page as needed:
% \ifCLASSOPTIONpeerreview
% \begin{center} \bfseries EDICS Category: 3-BBND \end{center}
% \fi
%
% For peerreview papers, this IEEEtran command inserts a page break and
% creates the second title. It will be ignored for other modes.

\IEEEpeerreviewmaketitle
	
	%%%%%%%%%%%%%%%%%%%%%%%%%%%%%%%%%%%%%%%%%%%%%%%%%%%%%%%%%%%%%%%%%%%%%%%%%%%%%%%%
	\section{INTRODUCTION}
	
	\IEEEPARstart{W}ITH the development of 3D sensors, the 3D point cloud completion task has attracted the attention of more and more researchers. The completion task plays an important role in subsequent robot navigation\cite{Engel2014LSDSLAMLD}, scene understanding\cite{Hou20193DSIS3S}, AR reconstruction\cite{Boud1999VirtualRA} and other applications. However, due to sparsity and the pose variety of 3D sensors, completion under factors such as severe occlusion and geometry deformation still remains largely unexplored.
	\begin{figure}[ht]
		\centering
		\framebox{\parbox{3.2in}{
				\centering

				\includegraphics[scale=0.28]{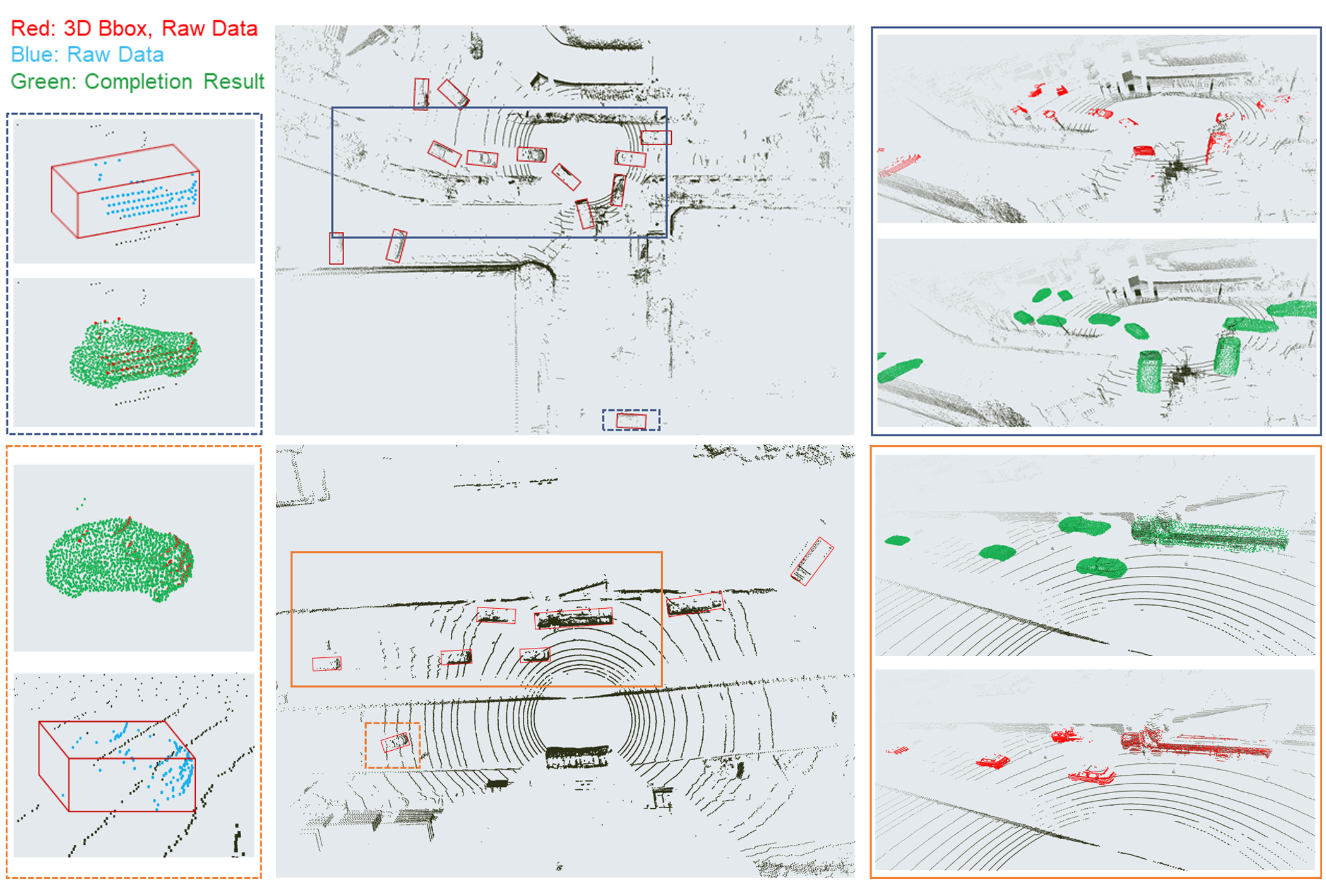}
		}}
		\caption{Visualized qualitative overview of completion results in self-driving scenes where the point cloud is quite sparse. We show the top view(middle) and the side view from a random perspective(right) for crossroads and straight roads, and select two vehicles with occlusion and different geometry details in the perspective to show the completion result separately(left). More results can be found in the video https://www.youtube.com/watch?v=97JSXOkvuOI}
		\label{panda}
	\end{figure}
	Researchers have given a variety of effective methods based on voxel\cite{Dai2017ShapeCU,Han2017HighResolutionSC,Xie2020GRNetGR}, point cloud\cite{Wen2020PMPNetPC,Wen2020PointCC,Yin2018P2PNET} and implicit field\cite{Gao2020LearningDT,Deng2020DeformedIF}. 
	
	The voxel-based completion methods first decompose the space into voxel sets and infer the information of each voxel or its corners. Such methods take into account the sparsity of the input point cloud, and can effectively reduce the storage space requirements. But at the same time, they are also subject to a resolution limitation, often disabling the recovery of complex surface information. Similarly, the implicit field method usually relies on the voxel structure to obtain signed distance field(SDF) data\cite{Mescheder2019OccupancyNL}, and lacks the ability to preserve details effectively. In addition, SDF methods often spend a lot of time sampling points around the object, taking much unnecessary time during training and inference\cite{Thai20203DRO}. Improving on the aforementioned methods, methods based on point clouds directly, such as PCN\cite{Yuan2018PCNPC} and TopNet\cite{Tchapmi2019TopNetSP}, have advanced the state-of-the-art on the point cloud completion task. These methods generally use an encoder-decoder structure to extract the overall shape information from the input data, and use this global shape information to generate the final complete point cloud. 
		\begin{figure*}[ht]
		\centering
		\vspace{1em}
		\framebox{\parbox{6.8in}{
				\includegraphics[scale=0.51]{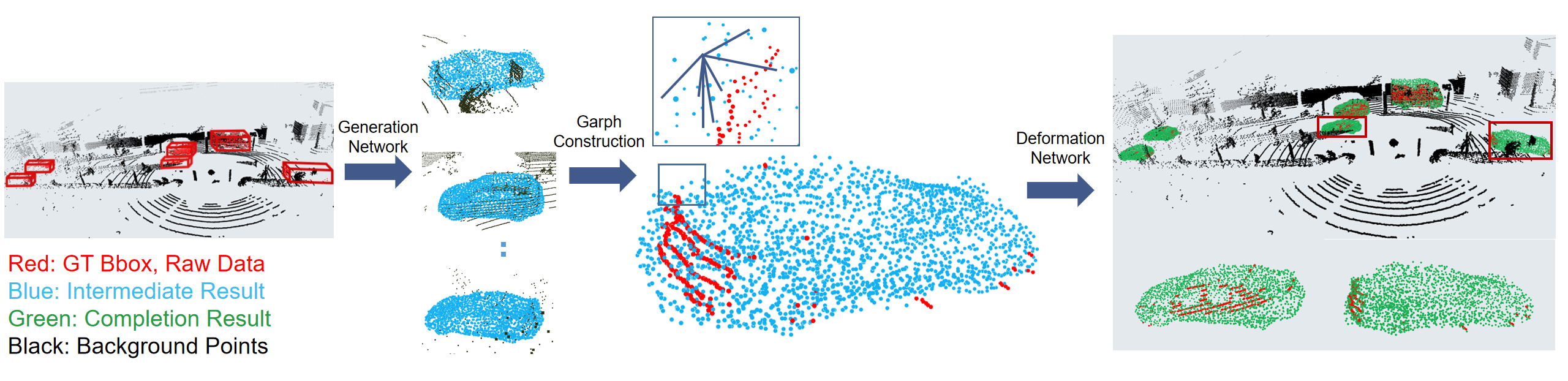}
		}}
		
		\caption{Illustration of the network. The input point cloud is processed by a point generation network to get an intermediate result and then refined by our deformation network. It can be seen that there are some deviations in scale and details in the intermediate point cloud. Through our deformation algorithm, such weaknesses are greatly improved in the final result. }
		\label{framework}
		\vspace{-1em}
	\end{figure*}
	While it is conceptually simple to borrow networks based directly on point clouds from other basic tasks, it is not obvious which network architectures will yield robust and refined performance for completion. In particular, the simple generalization-conjecture completion method often tends to synthesize the generated objects into the average value of a certain kind of objects, which unfortunately lacks much local information. In our opinion, this problem mainly stems from under-utilization of the input data.
	
	To go deeper by leveraging structure and local information of sparse 3D data, we look to traditional graphics problems for help. It is not difficult to find that if we obtain the edge information between point clouds, we can regard the point cloud completion problem as a mesh deformation problem with control points\cite{SorkineHornung2004LeastsquaresM}. Among the points, the input point cloud is the controlling points whose coordinates have been determined, and the other points to be completed are the point clouds to be restored, which have lost the coordinates information and only have a Laplacian coordinate\cite{Field1988LaplacianSA} relationship. Different from deep-learning-based mesh deformation tasks\cite{Wang2018Pixel2MeshG3,Shen2020InteractiveAO,Gkioxari2019MeshR}, we do have some reliable input points that can be made use of, and thus do not need to generate the whole object from a random shape. Therefore, we propose a graph-guided deformation network, which decomposes the point cloud completion problem into two tasks: edge generation and deformation. 
	
	In order to generate the edges, we first make a simple preliminary completion, that is, use the common encoder-decoder structure to generate the approximate point cloud shape. On this basis, we assume that there are unidirectional connected edges between adjacent points and construct the association graph. Here, we get the controlling points from the original input point cloud, the coordinates of the points to be restored after initialization, and the edge connection relationship between the point clouds. After that, we use a Graph Convolutional Network(GCN) to simulate the traditional Laplace optimization process\cite{SorkineHornung2005LaplacianMP, Field1988LaplacianSA} and get the final output. With experiments on ShapeNet\cite{Chang2015ShapeNetAI}, KITTI\cite{Geiger2012AreWR,Geiger2013VisionMR} and a self-collected dataset, we prove that our method is simple but effective. Our key contributions are as follows:
	\begin{itemize}
		\item[1.] Proposing a simple deformation network for point cloud completion and obtaining the state-of-the-art quantitative result. Our structure is easy to migrate and can be added to any existing completion models to refine the generated outputs.
		\item[2.] Using the idea of controlling points to fuse the input and output information, which fully exploits reliable input to get richer details.
		\item[3.] Creatively transforming the point cloud completion problem into a traditional mesh deformation task, and proposing an effective edge linking and deformation method to reduce the gap between the point cloud completion and the reconstruction problem.
	\end{itemize}
	\section{Related Work}
	3D point cloud completion has long been a hot topic for researchers, especially with the development of deep learning techniques. In this section, we focus on deep-learning-guided point completion methods, and extensively discuss different completion-related topics.
	\subsection{Shape Encoding}
	At present, almost all the work on point cloud completion is based on an encoder-decoder structure. The input data is firstly processed to extract the shape information, and then the completed point cloud is generated based on the global shape code. Researchers work on different input data types, such as point cloud\cite{Wen2020PMPNetPC,Yin2018P2PNET,Nie2020SkeletonbridgedPC,Huang2020PFNetPF,Wang2020CascadedRN}, voxels\cite{Dai2017ShapeCU,Han2017HighResolutionSC,Stutz2018Learning3S,Xie2020GRNetGR}, and SDF\cite{Gao2020LearningDT,Thai20203DRO,Liu2020DISTRD}, but they all follow a similar generation-inference process. For example, 3D-EPN\cite{Dai2017ShapeCU}, the typical voxel-based method, processes coordinate information and SDF information at the voxel level and add a 3D classification branch to enhance the shape code. In implicit methods, DeepSDF\cite{Park2019DeepSDFLC} explicitly uses simple Multi-Layer Perceptron(MLP) to generate a separate shape code for each class of object, so as to encode the common information of the class and infer the SDF values. Similarly, point-based methods employ a modified PointNet\cite{Qi2017PointNetDL,Qi2017PointNetDH} framework to encode the shape information of points and operate different encoding based on the shape code. 
	
	To encode more detailed information into the global shape, researchers have frequently explored how to generate a better shape code for the following generation operations. In addition to adding an auxiliary task, there have also been attempts centered on the design of the encoder. For example, \cite{Nie2020SkeletonbridgedPC,Huang2020PFNetPF} utilize a multi-resolution encoder to model raw point clouds at different scales. The ultimate goal of these methods is to generate a shape code that can fully represent the object category and describe the details of the object.
	\subsection{Shape Completion}
	While the pioneering researchers employed only the global shape code to generate the final point cloud through stacked MLP layers\cite{Tchapmi2019TopNetSP,Yuan2018PCNPC} or Generated Adversarial Networks\cite{Achlioptas2018LearningRA,Shu20193DPC,Sarmad2019RLGANNetAR}. It is obvious that relying solely on the shape code for completion will lose significant details of the object. We therefore propose to combine global shape information with local details to get a refined point cloud.
	
	A typical upgrade to extract more local information is to add the skip connection structure such as U-Net in 2D image perception\cite{Ronneberger2015UNetCN}. Considering that the lower features in the encoder contain more global information, while the upper features contain more local details, the simplest method of information fusion is to directly fuse the upper features into the decoder. It is easy to get the features of the corresponding positions by using voxel-based decoders and a 3D CNN and directly splicing the features by imitating the two-dimensional method. But voxel-based methods generally use binary representation, that is, to denote a voxel as occupied or not\cite{Mescheder2019OccupancyNL}, and use the midpoint coordinates to represent all internal points. Obviously, low resolution will lead to the loss of detailed information, while too intensive space division will require many computing resources. Thus, voxel-based methods seem unable to maintain efficiency and accuracy at the same time. GRNet\cite{Xie2020GRNetGR} tries to reduce the loss of information by encoding voxel corners into the voxel feature, but it is still subject to the resolution. Implicit-field-based methods often rely on the voxels as an intermediate representation, or rely on the generated point clouds. Thus, their accuracy is often affected by the intermediate results. Point-based methods do not limit the size and resolution of the input object. However, the point cloud is often orderless, so researchers can not employ the same skip layer as voxel-based methods do. Thus, SA-Net\cite{Wen2020PointCC} turns to the attention strategy for help, and uses the skip attention to learn a weighted addition at the feature level.
	
	Such skip structures can recover the high-level information lost in the convolution process. But there is a certain difference between the point cloud completion task and the traditional two-dimensional image perception, which is that the input point clouds are usually reliable and accurate, and relying on skip addition alone does not make full use of the input information. Therefore, some researchers have considered directly using the input information to fuse the input with the intermediate generated results\cite{Wen2020PMPNetPC,Nie2020SkeletonbridgedPC}, such as using the attention strategy or a gated structure to fuse the point from the feature level. However, these methods usually separate the input and generated information and, thus, can not use the input information to guide the generation of output results while keeping the input position unchanged.
	\subsection{Mesh Deformation}
	Another direction related to our task is mesh deformation\cite{Gao2020LearningDT, Wang2018Pixel2MeshG3, Shen2020InteractiveAO, Groueix2018APA, Gao2019DeepSplineDR}, which is a popular branch of 3D reconstruction. To get the reconstructed model of a specific object, researchers usually choose an initial reference mesh and deform it step by step for the desired results. For example, Pixel2Mesh\cite{Wang2018Pixel2MeshG3} takes an ellipsoid with fixed vertices and edges as the reference mesh, and outputs a deformed object corresponding to the input image. To avoid the shape limitation caused by the fixed initialization mesh, AtlasNet\cite{Groueix2018AtlasNetAP} first scatters the intermediate points according to the shape code, and then generates the final mesh by learning super-parameters. Mesh-RCNN\cite{Gkioxari2019MeshR} combines the mesh deformation with occupancy prediction, obtaining a more accurate intermediate result by predicting the occupancy information of voxels and deforming the mesh based on prediction. Finally, DEFTET\cite{Gao2020LearningDT} further employs a tetrahedron representation so that the surface can be deformed in a more casual and convenient way.
	
	The mesh deformation methods are very enlightening to us. Using known structure information as the initial step, we can deform the intermediate result to get a smoother and more detailed result. But popular 3D reconstruction tasks usually get mesh results directly from a simple monocular image. Differently, we already have a reliable input information from the beginning, which can be used to better guide the subsequent deformation.
	\section{Method}
	\begin{figure*}[h]
	    \vspace{1em}
		\centering
		\framebox{\parbox{6.4in}{
				\includegraphics[scale=0.48]{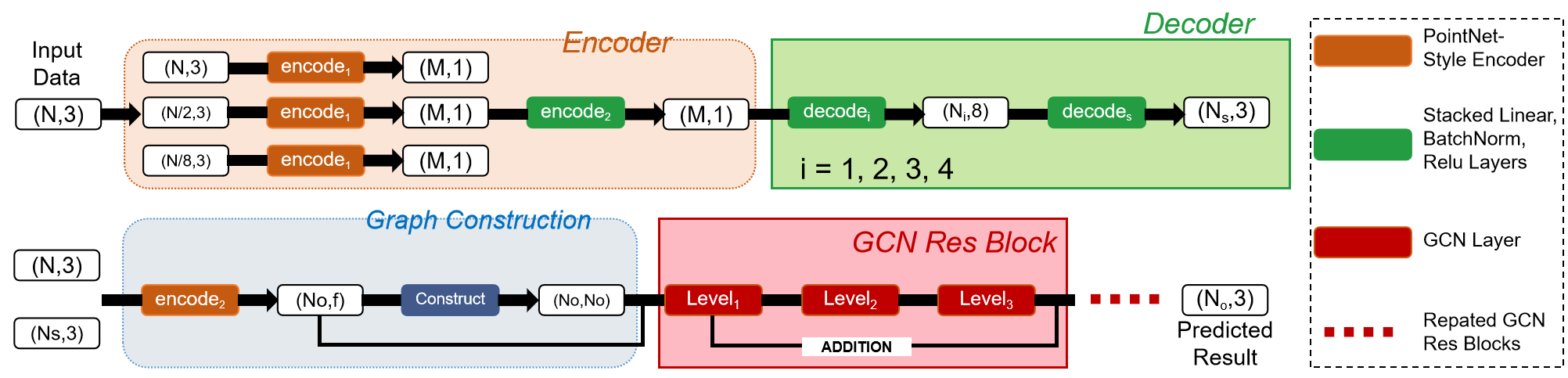}
		}}
		\caption{Details of our network. Upper: Shape-Preserved Generation Network. Lower: Deformation Network.}
		\label{network}
		\vspace{-1em}
	\end{figure*}
	Our network can be divided into two parts. The first is a shape-reserved generation network, as is illustrated in figure \ref{network}. In this section, we use a simple encoder-decoder structure to recover an approximately uniform point cloud $P_s$, which is used as the initialization result of the second module, a deformation network implemented by a GCN. In the deformation network, we first calculate the correlation graph $A$ of the point cloud, and then use the GCN to simulate the Laplacian optimization process, and optimize the point cloud to the final output $P_o$.
	\subsection{Shape-Preserved Generation Network}
	Suppose that the size of the input point cloud $P$ is $N$, and the size of the final output point cloud $P_o$ is $N_o$, where $N$ and $N_o$ can be different. The purpose of the generation network is to learn the overall shape information contained in $P$ and infer the intermediate result $P_s$ of size $N_s$. 
	
	In order to achieve this goal, we use a multi-resolution encoder structure, which is designed for detailed structure information encoding. We first apply furthest point sampling(FPS) to $P$, and obtain three point clouds with the size of $N$, $\frac{N}{2}$, and $\frac{N}{8}$. After that, MLP layers and global pooling structures are applied seperately to each point cloud to obtain three groups of features $f_1, f_2,$ and $f_3$ with size $(B, M, 1)$. We formulate this process as
	\begin{equation}
	f_i = concat(g \circ h_0(P), g \circ h_1 \circ h_0(P_i)), i = 1, 2, 3.
	\end{equation}
	where $h_0$ and $h_1$ are the MLP layers, and $g$ denotes the global pooling layer. The three sets of features are concatenated and passed through another set of MLP layers to obtain the final result $f_{final}$ of the same size $(B,M,1)$. We suppose that $f_{final}$ should contain the shape information of the input point cloud at different scales, which can summarize its main categories and shape structures.
	
	Considering that the main purpose of the generation network is not to obtain a final fine result, we select a simple tree structure for the decoder, following\cite{Tchapmi2019TopNetSP}. To get the final point cloud $P_s$ from the shape code $f_{final}$, we first use sequential MLP layers to change $f_{final}$ into shape $(B, 8, C_0)$, and then split the feature into eight different features of size $(B, C_0)$. For each generated feature, we iteratively pass it through the MLP layers and split it into eight features for a final result, which is reshaped to shape $(B, N_s, 3)$.
	\subsection{Graph Construction}
	After the simple generation network, we get a point cloud $P_s$ to be optimized. In order to obtain reliable control points, we perform the FPS operation again on the original input point cloud to obtain a point cloud $P_c$ of size $N_c$, where $N_c + N_s = N_o$. We fuse $P_s$ and $P_c$ directly for an intermediate result $P_g$, which already contains the trusted point $P_c$, as the controlling point, and the supporting point $P_s$ that needs to be optimized. 
	
	To further utilize the input information, we integrate the input data to $P_g$ through a modified PointNet\cite{Qi2017PointNetDL} encoder. For each ball radius $r, 2r, 4r$, we group $M$ points around each point within the ball. After the grouping operation, we operate feature extraction for each radius by stacked Conv1D, Relu and BatchNorm layers, and concatenate the three features together. Therefore, for every $p_i$ in $P_g$, we maintain another feature $F(p_i; P)$, where P is the original input point cloud.
	
	Due to the lack of specific structural information of objects, we need to create one-way or two-way edges between points for information transmission within the point cloud. Considering our ultimate goal, that is, to optimize the shape of an object and obtain detailed information by using the deformation network, we focus on establishing a local graph for better local information. Therefore, we perform a $k-NN$ operation on each point $p_i$ in $P_g$ to establish a one-way edge for the nearest $k$ points, and to establish a Laplacian coordinates map $A$ for $P_g$. In fact, considering that the number of adjacent points of each point is fixed to $k$, the Laplacian coordinates are the product of the adjacency table of points and the average coefficient $\frac{1}{k}$. Note that all the operations we mention here, including FPS, feature re-generation and graph construction, rely on the result of the previous module, and are carried out automatically every forward pass. Also, the loss of the whole network can be freely passed through such operations during the whole training process.
	\subsection{Graph Deformation Network}
	Now that we have obtained all preliminary information, our task has turned to  optimizing the position of $P_s$ by using the coordinates information of reliable points $P_c$ and the adjacent relationship $A$. It is not difficult to find that this optimization process is very consistent with the theory of the GCN, which also makes use of Laplacian information to refine adjacent nodes. Therefore, we adopt the G-ResNet\cite{Wang2018Pixel2MeshG3} structure, which is designed especially for 3D deformation. 
	
	We first process the input feature $f_i$ through a GCN layer, where
	\vspace{-0.5em}
	\begin{equation}
	f_i = concat(F(p_i; P), f_{final}, p_i, l_i),
	\end{equation}
	\begin{equation}
	p_j = GCN(f_1, f_2, ..., f_{N_g}; A).
	\end{equation}
	Here, $l_i$ is a one hot label, indicating whether the point is a controlling point. The GCN layer uses the Laplacian matrix $A$ to modify the weights of adjacent points and get a final result, which is similar to the attention module. In our network, we follow Pixel2Mesh\cite{Wang2018Pixel2MeshG3}, and arrange the GCN layers in a ResNet style, adding the outputs of two subsequent GCN layers together every two layers as a res-block. Different from \cite{Wang2018Pixel2MeshG3}, we re-compute the adjacent matrix every forward pass according to the result of the shape-preserved generation network and do not require a pre-defined mesh to refer to.
	\subsection{Loss Function}
	Our training losses have three parts. The first is the reconstruction loss that encourages the output to be the same as the ground truth. Second, we further design a match loss to ensure that the controlling points are not changed during the training process. Last, since we have established virtual edges between points, we add a shape-preserving loss following the mesh reconstruction tasks to limit the smoothness of the output along edges. 
	\subsubsection{Reconstruction Loss}
	We select the popular bi-directional Chamfer distance(CD) loss as the reconstruction loss, where
	\begin{equation}
	L_{CD}(X, Y) = \frac{1}{N_0}\sum_{x \in X}\min_{y \in Y}||x - y||^2_2 + \frac{1}{N_1}\sum_{y \in Y}\min_{x \in X}||y - x||^2_2.
	\end{equation}
	
	The CD loss calculates the nearest distance from each point to another point cloud. By calculating the bi-directional CD loss, we are able to limit the overall shape of the output point cloud. In our experiments, we calculate the CD loss twice: for the output of the generation network and the final result:
	\vspace{-0.5em}
	\begin{equation}
	L_{recons} = L_{CD}(P_s, P{gt}) + \lambda L_{CD}(P_o, P_{gt}).
	\end{equation}
	\vspace{-1em}
    \begin{table*}[ht]
	\centering
	\vspace{1em}
	\setlength{\abovecaptionskip}{0pt}%    
	\setlength{\belowcaptionskip}{0pt}%
	\caption{Comparison on ShapeNet Dataset in terms of per-point Chamfer Distance $\times 10^4$(Lower is Better)}
	\begin{tabular}{c|c|cccccccc}
		\toprule
		Methods & Average & Plane & Cabinet & Car & Chair & Lamp & Couch & Table & Watercraft\\
		\midrule
		AtlasNet\cite{Groueix2018AtlasNetAP} & 17.69 & 10.37 & 23.4 & 13.41& 24.16 & 20.24 & 20.82& 17.52 & 11.62\\
		PCN\cite{Yuan2018PCNPC} & 14.72 &8.09 & 18.32 & 10.53 & 19.33 & 18.52 & 16.44 & 16.34 & 10.21\\
		FoldingNet\cite{Yang2018FoldingNetPC} & 16.48 & 11.18 &20.15 &13.25 &21.48 &18.19 &19.09 &17.8 &10.69\\
		TopNet\cite{Tchapmi2019TopNetSP} & 9.72 &5.5 &12.02 &8.9 &12.56 &9.54 &12.2 &9.57& 7.51\\
		SA-Net\cite{Wen2020PointCC} & 7.74 &\bf{2.18} &9.11 &\bf{5.56} &8.94 &9.98 &7.83 &9.94 &7.23\\
		\midrule
		Ours & \bf{6.02} & 2.50 & \bf{8.35} & 6.34 & \bf{6.58} & \bf{6.19} & \bf{7.69} & \bf{5.63} & \bf{4.84}\\
		\bottomrule
	\end{tabular}
	\label{shapenet-cd}
	\vspace{-2em}
    \end{table*}	
	\subsubsection{Matching Loss}
	We point out that the order of points within the point cloud will not be changed during the deformation operation. Therefore, we record the order of the whole point cloud after graph construction, and add a matching loss for controlling points. Suppose the controlling points turn to $P_c'$ after deformation. We use a typical smooth-l1(Huber) loss with $\sigma = 2$ as $L_{match}$.
	\subsubsection{Shape-Preserving Loss}
	The GCN has the ability to smooth nearby points and avoid sharp changes. In our realization, we choose the averaged Laplacian coordinates, named umbrella coordinates, to calculate the weights or nearby points. However, such a weighting scheme may lead to over-smoothing, which has been proved in traditional mesh smoothing tasks. Thus, we further add a shape-preserving loss to avoid such a result.
	
	One popular method to preserve the shape while smoothing the surface is to utilize the Taubin\cite{Taubin1995ASP} weighting scheme. However, such a weighting method is too complicated for our network, especially when we have to calculate the virtual edge and weighting graph every forward pass. Thus, we add a scale-dependent umbrella Laplacian loss\cite{Fujiwara1995EigenvaluesOL,Shen2020InteractiveAO} in the loss function, which guide the network in a back-propogation manner. The loss can be formulated as
	\vspace{-0.5em}
	\begin{equation}
	e = \sum_{x \in P_g, y \in A(x)} || x - y||_2,
	\end{equation}
	\begin{equation}
	L_{shape} = ||\sum_{x \in P_g} \sum_{y \in A(x)} \frac{2(x - y)}{e  ||x - y||_2} ||_2.
	\end{equation}
	
	Here x denotes all points in $P_g$, and y is the points adjacent to x in the Laplacian table $A$. This item ensures that the Laplacian coordinates of each point do not change much before or after the GCN process, thus ensuring the local shape of the whole point cloud.
	\vspace{0.3em}
	\subsubsection{Overall Loss}
	The overall loss is the weighted sum of the above three loss items:
	\begin{equation}
	L = 1000 * L_{recons} + \alpha L_{match} + \beta L_{shape}.
	\end{equation}
	
	During training, we adjust the weighting terms $\alpha$ and $\beta$ along with the training epochs.
	\section{Experiments}
	\subsection{Training Details}
	In the experiments, we set the weighting parameters in the loss functions as $\alpha = 1000$ and $\beta = 0.5$. The size of the input point cloud is 2048, which is also the size of the final result. We choose 2048 following the settings for PCN\cite{Yuan2018PCNPC}. In fact, the number of input data and output results do not have to be the same.  We further conduct a quantitative experiment and have an explanation of the number of input points in Section V, and we show that our network has a strong generalization ability.
	
	We select 256 as the number of controlling points in all experiments. Note that the number of controlling points influences the performance of the model. In synthetic datasets, a larger number may bring about much better results. However, we hope to show that only a small number of controlling points can help our network achieve an excellent refinement. We also analyze the performance of models with different controlling point sizes on the ShapeNet dataset in Section V.
	
	The network is trained from scratch on ShapeNet, and the same model is used for experiments on all datasets. We train the model end-to-end with the ADAM optimizer. For ShapeNet, we select batch size 32 and learning rate 0.01 for 70 epochs on one GTX 1080 Ti GPU. In the first 40 epochs, we set $\lambda = 0$ and focus only on the generation. We then set $\lambda = 3$ and refine the deformation network based on the intermediate output from the generation network.
	\subsection{Datasets}
	\subsubsection{ShapeNet}
	In order to ensure fairness, we conducte training and testing on the data provided by PCN\cite{Yuan2018PCNPC}. The training data are the partial point cloud obtained by projecting the ShapeNet\cite{Chang2015ShapeNetAI} model into 2D space, and back-projecting it into 3D space for a 2.5D partial model. The dataset includes eight kinds of objects: airplanes, cabinets, cars, chairs, lamps, sofas, tables and vessels. Note that the ground truth of \cite{Yuan2018PCNPC} is dense, and we uniformly sample 2048 points on the original mesh as the ground truth data. We take N=2048 as the number of input points by randomly sampling/upsampling the input data from \cite{Yuan2018PCNPC}. We also follow the train, val and test splits of \cite{Yuan2018PCNPC}, and choose the best model on the validation dataset for testing. Due to the randomness in point sampling, we conduct the experiment ten times and use the average value as our final result.
	
\begin{figure}[h]
	\vspace{1em}
	\centering
	\framebox{\parbox{3.2in}{
			\includegraphics[scale=0.36]{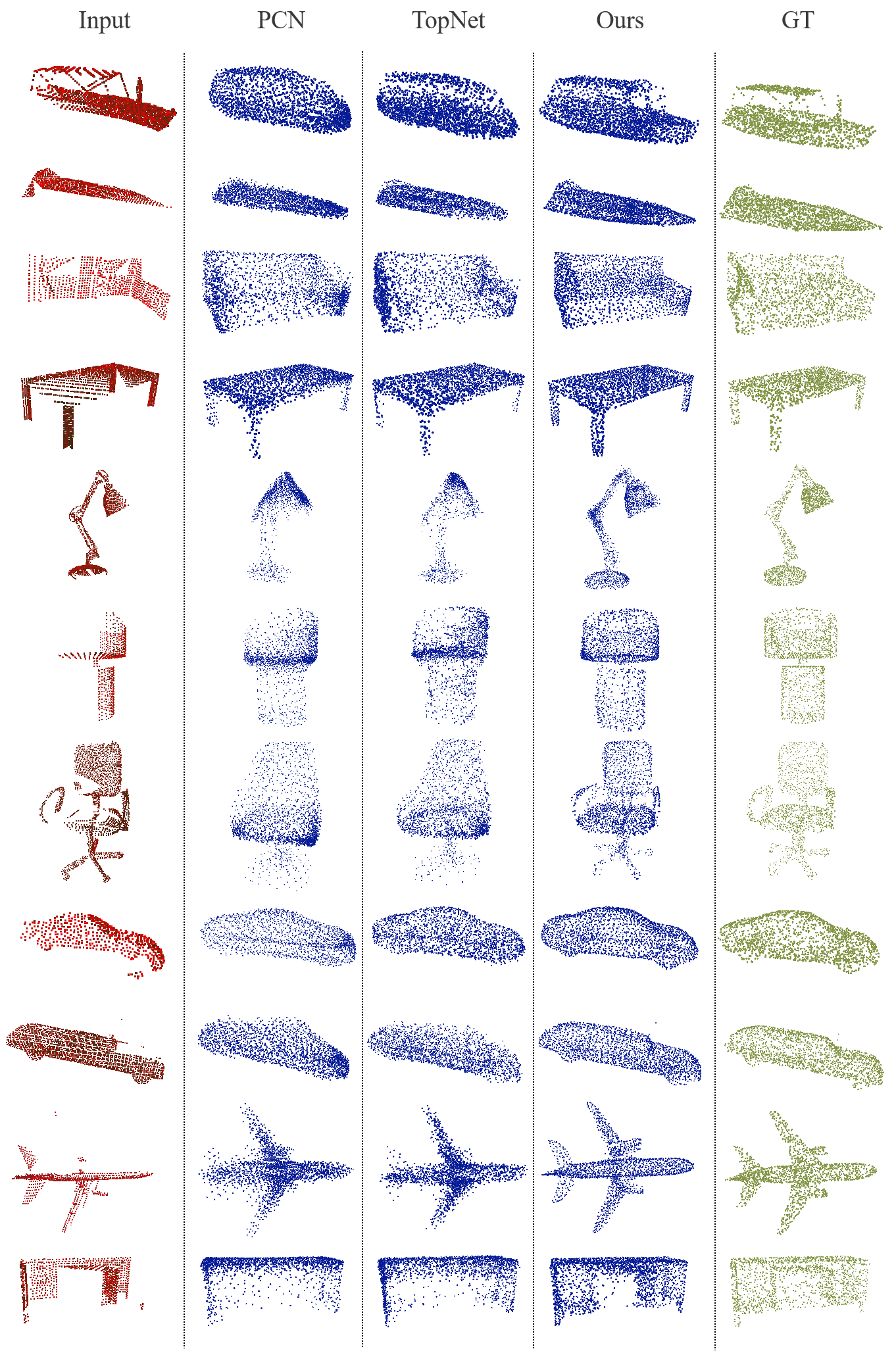}
	}}
	
	\caption{Visualized comparisons on ShapeNet Dataset. From left to right: Input Point Cloud, Results of PCN, Results of TopNet, Our Result, Ground-Truth.}
	\label{shapenet}
	\end{figure}

	Following SA-Net\cite{Wen2020PointCC}, we take the Chamfer Distance as the evaluation metric, and compare our results with AtlasNet\cite{Groueix2018AtlasNetAP}, PCN\cite{Yuan2018PCNPC}, FoldingNet\cite{Yang2018FoldingNetPC}, TopNet\cite{Tchapmi2019TopNetSP}, and SA-Net\cite{Wen2020PointCC} in Table \ref{shapenet-cd}. The results of the benchmarks are cited from the original paper of SA-Net\cite{Wen2020PointCC}. We also train the original version of PCN\cite{Yuan2018PCNPC} and TopNet\cite{Tchapmi2019TopNetSP} ourselves with the same training data, and provide visualized comparisons in Figure \ref{shapenet}. We can see that our method is more robust to irregular shapes. For example, in Figure \ref{shapenet}, line 5, we successfully recover the details of the irregular lamp, especially the handle position. And in Figure \ref{shapenet}, line 1, we recover the cavity of the watercraft cabin, while other methods tend to recover it as a solid object or a car of a similar shape. Our method also can preserve more details than the benchmarks. Through quantitative and qualitative comparisons, we prove the superiority of our method in point cloud completion as it can provide smoother and more realistic results, especially in the face of complex objects.

	\subsubsection{KITTI}
\begin{figure}
		\centering
		\vspace{1em}
		\framebox{\parbox{3.2in}{
				\centering
				\includegraphics[scale=0.27]{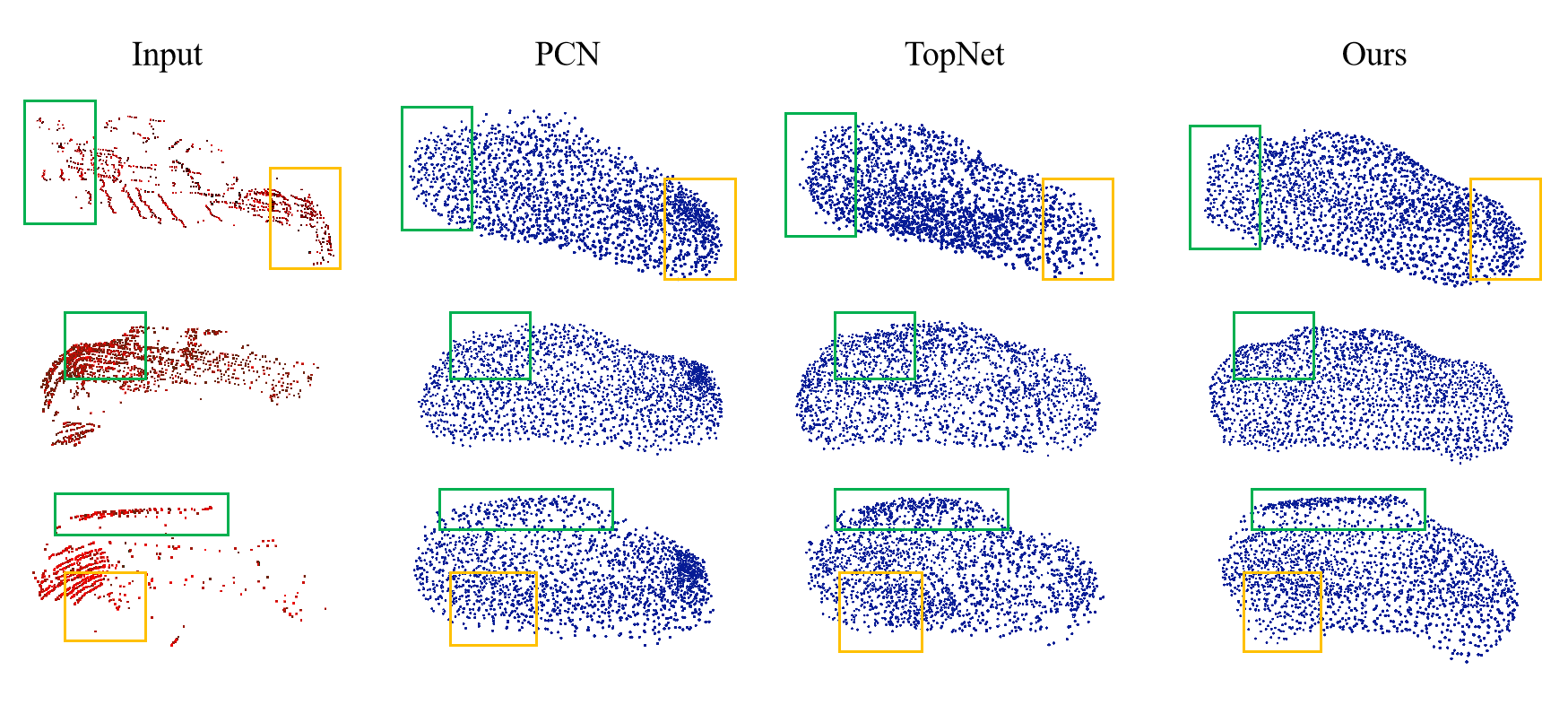}\
				\includegraphics[scale=0.26]{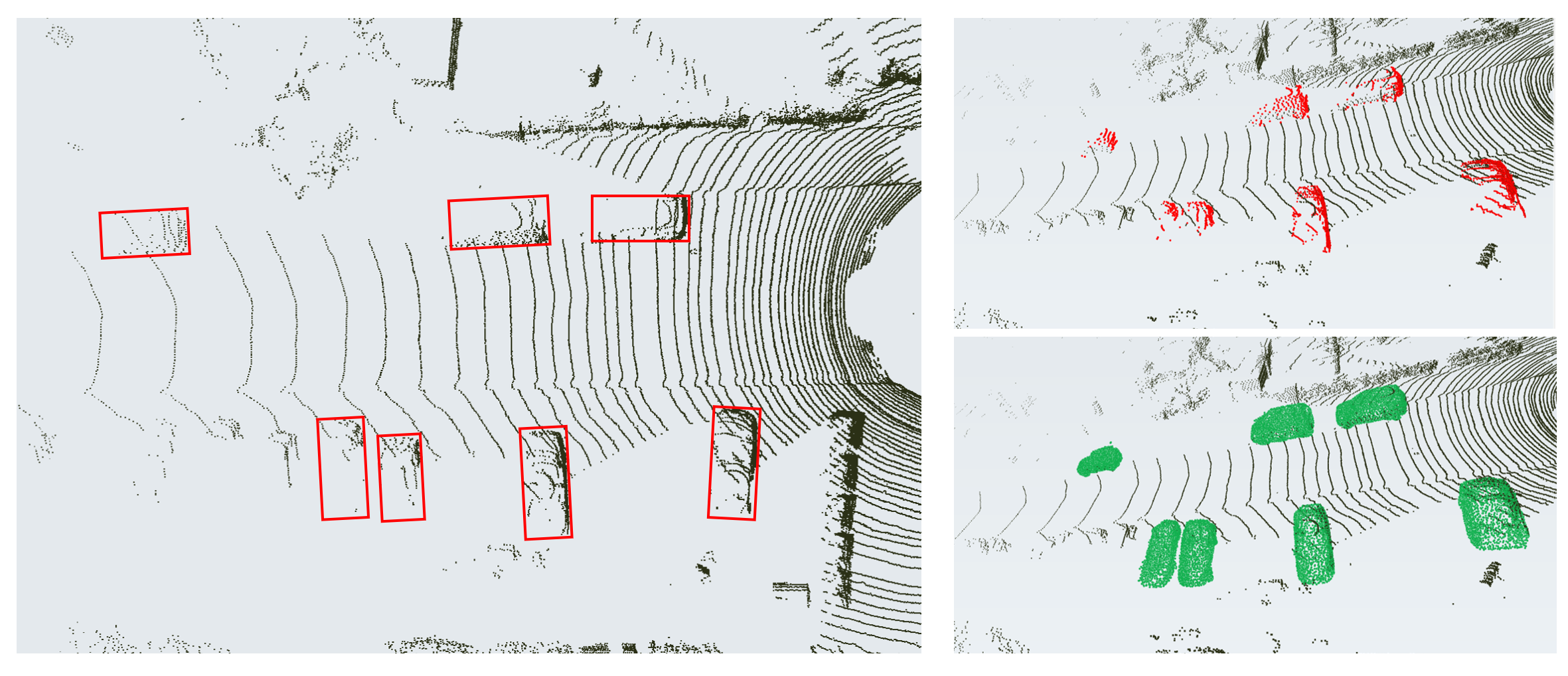}
		}}
		
		\caption{Visualized comparisons on KITTI Dataset. Our method still performs much better in preserving the details and recovering the shape.}
		\label{kitti}
	\end{figure}

	The KITTI dataset\cite{Geiger2013VisionMR} consists of several real-world self-driving scenes collected with lidar sensors. Though the data quality is high, KITTI does not provide a dense point cloud. Therefore, we make qualitive comparisons of the results of PCN\cite{Yuan2018PCNPC} and TopNet\cite{Tchapmi2019TopNetSP} on KITTI following \cite{Wen2020PointCC}. We only conduct point cloud completion on cars and vans. Since the number of points of each scan is not fixed, we randomly select 2048 points for each object as the input. As is shown in Figure \ref{kitti}, we frame out some obvious details, such as the front of the car, the roof and the tires. Obviously, our method can better restore the shape and subtle changes of the vehicle, while maintaining the integrity and smoothness of the overall model. 
	
	\subsubsection{Pandar40 Dataset}
	This part of the experiment is designed to assess the robustness of our approach. In practice, we collect raw point cloud data utilizing lidar sensors, which is much more sparse and noisy. The results suggest that our network is capable of handling migration and completion in real-world situations, rather than being only applicable to limited scenes.
	Here we introduce our self-collected Pandar40 dataset. We set up a Pandar40 Lidar, a 40-Channel Mechanical LiDAR from HESAI with 200m extended measurement range, 20\% reflectivity, 23$^{\circ}$ vertical field of view and 0.33$^{\circ}$ minimum vertical resolution, on a converted drone, and collected data of sunny/rainy days and nights on Qianhai Avenue of Shenzhen, China. Similar to KITTI\cite{Geiger2013VisionMR}, we label the 3D bounding boxes of car, van and truck manually. We still use the best completion model on ShapeNet with no fine-tune to finish the completion task. As can be seen in Figure \ref{panda}, our model manages to complete cars and trucks even though the point clouds are sparse, which proves that our model has strong migration ability and is quite robust.
	\subsection{Application: Point Cloud Completion for Classification}
	To test whether our completion models work for certain application tasks, we follow \cite{Sarmad2019RLGANNetAR} and \cite{Wang2020CascadedRN}, testing our network on the classfication task. We train a PointNet\cite{Qi2017PointNetDL} on the training data the same as our completion task. For testing, we use PCN, TopNet and our network to process the partial input in the test dataset, and input the completed point cloud into the trained classification network. For reference, we also take the ground truth point cloud as the input data and test the accuracy as the upper bound. We show in Table \ref{class} that our method outperforms the other two in the classification task, which demonstrates that our work has its practical value.
	
	\begin{table}[ht]
		\centering
		\vspace{-1em}
		\setlength{\abovecaptionskip}{0pt}%    
		\setlength{\belowcaptionskip}{0pt}%
		\caption{Comparison of Classification Accuracy(\%)}
		\begin{tabular}{c|c|c|c}
			\toprule 
			PCN\cite{Yuan2018PCNPC} & TopNet\cite{Tchapmi2019TopNetSP} & Ours & GroundTruth\\
			\midrule
			90.00\% & 90.42\% & 94.58\% & 96.33\%\\
			\bottomrule
		\end{tabular}
		\label{class}
	\end{table}
    \vspace{-1.5em}
	\section{Ablation Study}
	Apart from the experiments to validate our approach in comparison with the state-of-the-art methods on point cloud completion, we also conduct ablation experiments on the effects of important hyper-parameters of our model.
	\subsection{Effect of Deformation Network}
\begin{figure}[h]
		\centering
		\framebox{\parbox{3in}{
				\centering
				\includegraphics[scale=0.3]{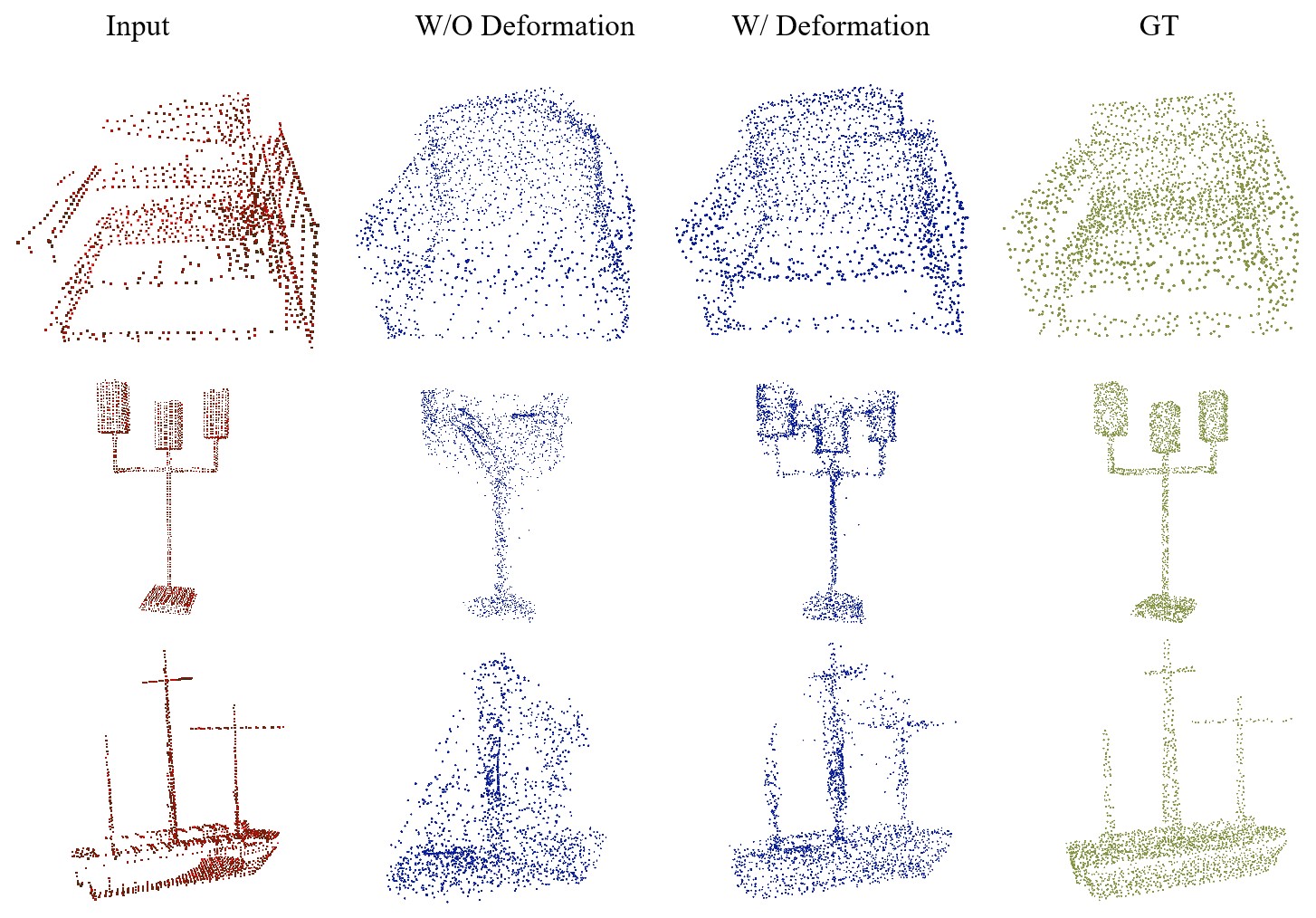}
		}}
		\caption{Comparison between model with and without deformation layers.}
		\label{deform}
	\end{figure}

	In order to verify the function of the deformation network, we omit the deformation layers and only train the generation network on the ShapeNet Dataset. The same optimizer, learning rate and batch size are used in the training. We use only CD loss and train for 50 epochs, and we select the best model in the validation set for comparison.
	\begin{table*}[h]
	    \vspace{1em}
		\setlength{\abovecaptionskip}{0pt}%    
		\setlength{\belowcaptionskip}{0pt}%
		\centering
		\caption{Results with and Without Deformation Layers on ShapeNet Dataset(Chamfer $\times 10^4)$}
		\begin{tabular}{c|c|cccccccc}
			\toprule
			Methods & Average & Plane & Cabinet & Car & Chair & Lamp & Couch & Table & Watercraft\\
			\midrule
			W/ Deformation & \bf{6.02} & \bf{2.50} & \bf{8.35} & \bf{6.34} & \bf{6.58} & \bf{6.19} & \bf{7.69} & \bf{5.63} & \bf{4.84}\\
			W/O Deformation & 7.56 & 3.11 & 9.54 & 6.59 & 8.61 & 9.70 & 8.62 & 7.36 & 6.85\\
			\bottomrule
		\end{tabular}
	\vspace{-2em}
		\label{deform-cd}
	\end{table*}
	We also select typical cases where adding the deformation network greatly improves the final result, and we present the visualized result in Figure \ref{deform}. We can see that the deformation layers work well especiall in complex cases, where a simple point generation network cannot learn complex structural changes. The spare controlling points sampled from the original input effectively guide the follow-up completion work and make the structure of the completion result clearer. This also explains our excellent results on the classification of watercraft and lamp in Table \ref{deform-cd}.
	\vspace{-0.5em}
	\vspace{-1.5em}
	\subsection{Effect of Number of Controlling Points}
	It is obvious that more controlling points will help the network preserve more details of the input data. Yet it will also limit the ability of the model to infer missing parts of a complete object. We choose 256 in our experiment to better show that our deformation layer can infer abundant information through few controlling points, though it is true that 256 is not the best choice on the ShapeNet dataset.
	
	To make it clear how controlling points affect the performance of our network, we conduct an experiment on the number of controlling points on the ShapeNet dataset. We select the number of controlling points $N_c = 128, 256, 512, 768$, and keep all other parameters the same. We also keep the same pre-trained generation network and fix the parameters, and we only re-train the deformation layers to avoid other effects. Here, we do not use the bi-directional CD loss, but list the details of both directions to better show the performance. Thus, we denote the deformed point cloud and the ground truth as D and G, and use $CD(D,G)$ and $CD(G,D)$ to mark the difference.
	\begin{table}[h]
    	\vspace{-1em}
		\setlength{\abovecaptionskip}{0pt}%    
		\setlength{\belowcaptionskip}{0pt}%
		\centering
		\caption{Experiments on Controlling Points Number.}
		\begin{tabular}{c|c|c|c|c}
			\toprule
			Loss Direction & 128 & 256 & 512 & 1024 \\
			\midrule
			$CD(D,G) \times 10^4$ & 2.99 & 2.96 & 2.89 & \bf{2.85}  \\
			$CD(G,D) \times 10^4$ & 3.13 & 3.06 & \bf{3.05} & 3.09 \\
			\bottomrule
		\end{tabular}
		\label{control}
		\vspace{-1em}
	\end{table}
	Consistent with our estimation, when the number of controlling points increases, the generated results get better and more sufficient guidance, and thus tend to be similar to the real data. But at the same time, too many controlling points will lead to a failure to recover the missing part which is not covered by the input data, resulting in incomplete and uneven shape recovery. Therefore, the error from D to G decreases, but the error from G to D begins to rise after falling to a certain minumum.
    \vspace{-0.5em}
	\subsection{Effect of Number of Input Points}
	We have proved on KITTI and the self-collected dataset that our model also does well with sparse point clouds. However, we still wonder how it deals with different numbers of input points. Thus, we qualitively analyze the effect of the input point number on the ShapeNet dataset. Similar to the previous section, we conduct ten experiments for each case, and take the average value as the final result. 
	
	We use two strategies to sample N = 2048, 1024, 512, 256 points from the input point clouds. The first is to randomly select N points from the original point cloud, and put the sampled point clouds directly into the network. Another sampling method is to apply FPS to the input data for a set of more representative key points, and feed them to our network. The size of the output and ground-truth point clouds remain unchanged at 2048. Additionally, we want to figure out how much our network is influenced by the size of the input data. Therefore, we also try to lift the input points to 2048 by adding repeated points, and test the same two sampling strategies. To simplify the notation, we record two sampling methods as R and F, and use w., w/o lift to specify whether the input number is lifted to 2048.
	\begin{table}[h]
        \vspace{-1em}
		\setlength{\abovecaptionskip}{0pt}%    
		\setlength{\belowcaptionskip}{0pt}%
		\centering
		\caption{Experiments on Input Points Number.}
		\begin{tabular}{c|c|c|c|c}
			\toprule
			N & R w/o Lift & R w. Lift & F w/o Lift & F w. Lift \\
			\midrule
			2048 & 6.02 & * & \bf{5.99} & * \\
			1024 & 6.01 & 6.01 & \bf{6.00} & \bf{6.00}\\
			512 & 6.11 & 6.11 & 6.03 & \bf{6.02}\\
			256 & 6.48 & 6.51 & \bf{6.13} & 6.14\\
			\bottomrule
		\end{tabular}
		\vspace{-1em}
		\label{deform-N}
	\end{table}
	
	In Table \ref{deform-N}, we show that the number of input data does not need to be exactly the same as that of the output data, which is consistent with our claim in Section IV. What's more, we retain a state-of-the-art result when N equals 2048 or 1024. Considering the statement that almost all models in ShapeNet can be concluded by 1024 points\cite{Qi2017PointNetDH}, we think such a result is acceptable. Moreover, we find that the FPS method improves the result, especially when the number of input points is small. This result is consistent with that of the traditional mesh reconstruction method based on controlling points, which is that, when the input data contains more and more comprehensive structural information, the controlling points can effectively guide the reconstruction of objects and obtain more results with less data. 
\section{Conclusion}
In this paper, we propose a method to optimize the point completion task by simulating least square mesh deformation. By treating the input point cloud as a controlling point and using the popular generation network as an initializer, we make full use of the input information and obtain a smoother and more consistent result. Experiments on KITTI and a self-collected datasets show the robustness and scalability of our method, and the results on the classification task further prove that our method has practical value.
\bibliographystyle{IEEEtran}
\bibliography{root}
\end{document}